# ENHANCING ML MODEL INTERPRETABILITY: LEVERAGING FINE-TUNED LARGE LANGUAGE MODELS FOR BETTER UNDERSTANDING OF AI

*Completed Research Paper*

Jonas Bokstaller, University of Liechtenstein, Vaduz, Liechtenstein, jonas.bokstaller@uni.li

Julia Altheimer, University of Liechtenstein, Vaduz, Liechtenstein, julia.altheimer@uni.li

Julian Dormehl, Technical University of Munich, Munich, Germany, julian.dormehl@tum.de

Alina Buss, Technical University of Munich, Munich, Germany, alina.buss@tum.de

Jasper Wiltfang, Technical University of Munich, Munich, Germany, jasper.wiltfang@tum.de

Johannes Schneider, University of Liechtenstein, Vaduz, Liechtenstein, johannes.schneider@uni.li

Maximilian Röglinger, University of Bayreuth, Bayreuth, Germany, maximilian.roeglinger@fim-rc.de

## Abstract

*Across various sectors applications of eXplainableAI (XAI) gained momentum as the increasing black-boxedness of prevailing Machine Learning (ML) models became apparent. In parallel, Large Language Models (LLMs) significantly developed in their abilities to understand human language and complex patterns. By combining both, this paper presents a novel reference architecture for the interpretation of XAI through an interactive chatbot powered by a fine-tuned LLM. We instantiate the reference architecture in the context of State-of-Health (SoH) prediction for batteries and validate its design in multiple evaluation and demonstration rounds. The evaluation indicates that the implemented prototype enhances the human interpretability of ML, especially for users with less experience with XAI.*

## 1 Introduction

AI is seen as a transformative technology that is difficult to interpret (Berente et al., 2021). Explainable AI (XAI) aims to enhance the interpretability of Machine Learning (ML) predictions by providing explanations to the end user (Meske et al., 2022). One major goal of XAI is to enable humans to thoroughly understand and effectively interact with Artificial Intelligence (AI) models, which is highly relevant to change the development process from trial-and-error to be more targeted as well as to ensure the proper functioning of AI, e.g., by certification authorities (Meske et al., 2022). As such, XAI can help to mitigate safety risks. However, the status quo in interpretability research has been called largely unproductive (Räuker et al., 2023) and many open problems remain (Schneider, 2024). For example, little progress has been made, e.g., when it comes to interactive and more custom (or personalized) explanations (Schneider, 2024). At the same time, Large Language Models (LLMs) have recently shown remarkable interaction and customization capabilities. Only a few approaches have attempted to build systems combining XAI and LLMs (Gao et al., 2022; Slack et al., 2023; Nguyen et al., 2023). All of them do not tailor the model to the task but rather build around it, i.e., they tend to rely on rules, templates, and question banks rather than adjusting or augmenting the model itself. As such they yield promising but not yet convincing outcomes, when it comes to inferential capabilities, e.g., rather than just displaying importance values, connecting them with domain-specific knowledge in textual form. Our study aims to answer the Research Question (RQ): *"How can an LLM be leveraged to enhance existing XAI outputs for improved human understanding?"* One approach to achieving this is fine-tuning, which enables the model to grasp domain-specific nuances and generate explanations that are more accurate and contextually relevant. Unlike static rule-based methods, fine-tuning allows LLMs to move beyond generic patterns and provide richer, more insightful interpretations of importance values.

As data source to the LLM, various XAI libraries are available, each designed to provide post-hoc evaluations of ML models. The artifact we present is compatible with widely used XAI libraries; however, for this paper, we emphasize SHapley Additive exPlanations (SHAP) (Lundberg & Lee, 2017) due to its support for categorical input features. SHAP offers a fine-grained analysis of how model features contribute to the final ML prediction through a game-theoretic approach. Nevertheless, interpreting SHAP outputs requires domain-specific expertise in data science and statistical concepts. Despite its huge potential to increase interpretability, the comprehension of XAI techniques such as SHAP can be difficult for users who lack the knowledge to interpret SHAP outcomes, restricting the area of application to domain experts (Kaur et al., 2020). By leveraging SHAP values within the context of an interactive chatbot, users can engage in a dialogue-driven exploration of the model's predictions which has the potential to make the functionality of XAI techniques accessible to a larger user group. Therefore, we aim to bridge this gap by designing an interactive chatbot with XAI translation capabilities to facilitate the interpretation of ML models for human users. We follow an adapted version of the design science research (DSR) (Hevner et al., 2004; Peffers et al., 2007) approach, integrating the Framework for Evaluation in Design Science (FEDS) (Venable et al., 2016). We derive and formulate relevant Design Objectives (DOs) that we evaluate with expert interviews regarding their importance. Building upon those well-justified DOs, we iteratively develop and evaluate an artifact outlining the interaction of a fine-tuned LLM with SHAP to improve ML model interpretability. While doing so, we develop a fine-tuning process for the LLM consisting of three subsequent fine-tuning steps to increase its capability to comprehend, analyze, and respond to domain-specific user questions. We demonstrate that domain-specific fine-tuning results in a substantial improvement across all evaluated metrics. We also evaluate and demonstrate the real-world applicability of our artifact by implementing and demonstrating a prototype for a use case in the context of State of Health (SoH) prediction for rechargeable batteries. The use case is especially important not only for informative reasons but also for health and safety. An online survey concludes our research by confirming the benefits of our artifact to enhance the interpretability of ML models. Our artifact leverages an LLM, which can enhance pure numbers in the form of SHAP values with domain-specific context and ultimately lead to easy-to-understand information for the end user. This lowers the knowledge requirements for the end users making the interpretation of the model's prediction available to a wider audience.

## 2    Theoretical Background

### 2.1    Large language models

Large Language Models (LLMs) have evolved from early statistical language models and word embeddings using deep neural networks (Liu et al., 2017) to modern transformer-based architectures (Vaswani, 2017), significantly improving natural language understanding and generation. At the same time, the field of XAI evolved rapidly over the last years (Meske et al., 2022) though it is still under active research (Schneider, 2024). Initially, these models are trained on large-scale datasets for a broad understanding of language from raw text through unsupervised training techniques. While this broad knowledge base is essential for general language understanding, it is not always sufficient for specialized tasks or specific domains. Fine-tuning, on the other hand, is a subsequent process that adapts this generalized knowledge to particular applications. During fine-tuning, the model's weights are adjusted based on a domain-specific dataset, refining the model's understanding and improving its performance on specific tasks. For example, when fine-tuned on financial data, BloombergGPT showed significantly enhanced performance in financial tasks without losing its general capabilities (Wu et al., 2023).

Similarly, GitHub Copilot, fine-tuned for code generation, can generate code in multiple programming languages based on natural language descriptions (Nguyen & Nadi, 2022). These examples demonstrate how fine-tuning transfers knowledge from a broad linguistic context to more niche areas, optimizing performance without the need for full retraining from scratch. One notable approach is parameter-efficient fine-tuning (Hu et al., 2023) and more specifically, Low-Rank Adaptation (LoRa), which helps the prevailing model to transfer knowledge from the source tasks to the target task more effectively requiring only little memory and reducing issues such as catastrophic forgetting by fitting only a small number of parameters (Hu et al., 2022). Fine-tuning can be conducted either supervised or unsupervised. Unsupervised fine-tuning allows the model to discover the underlying patterns and structures related to the domain. It enhances the adaptability and robustness of domain-specific text generation. On the other hand, supervised fine-tuning utilizes structured data with explicit task-related annotations to guide the learning process. This aligns the model with the task objective and improves the accuracy of generated text outputs (Gunel et al., 2021). While the previous mentioned literature has primarily addressed the technical aspects of LLMs, research has expanded beyond technical concepts to include socio-technical perspectives (Chakraborty et al., 2017; Lehmann, 2022; Jussupow et al., 2020). These focus on the user's needs and interactions with AI systems.

## 2.2     eXplainable AI

XAI distinguishes between local and global explanations. Local explanations pertain to the prediction or decision for a specific instance in the dataset. Global explanations provide an overall understanding of the underlying model across the entire dataset. Secondly, XAI techniques are differentiated based on whether they provide post-hoc explanations or intrinsic explanations. Post-hoc explanations refer to explanations after the model has made a prediction. In contrast, intrinsic explanations comprise ML models that are inherently explainable such as decision trees or naïve Bayes classifiers (Arrieta et al., 2020; Brasse et al., 2023). A popular XAI method is SHAP, which represents a post-hoc explanation technique (Lundberg & Lee, 2017). Explanations are provided after a model has made a prediction. Originally coming from the field of cooperative game theory and adapted to the context of XAI to quantify the contributions of individual features in black box ML models, the contribution and therefore importance of single features to a prediction can be identified (Van den Broeck et al., 2022). By computing the effect of a feature on the model prediction, the SHAP values can be calculated and the model with its features becomes more comprehensible through the provided visualizations (Lundberg & Lee, 2017).

## 2.3     Related Work

Within screening the literature, we have identified four papers that integrate LLMs to enhance ML model interpretability. Gao et al. (2022) propose a chatbot explanation framework that can serve different purposes for the end user. As a use case, they implement a chatbot that interacts with an ML model for anomaly detection in a train ticket booking system. The chatbot is built with the Watson Assistant Service (IBM, 2024) and integrated into a productivity tool, i.e., a Slack app. Slack et al. (2023) introduce the *TalkToModel*. It is an interactive dialogue system that allows the user to get insights into ML models through natural language conversations. The user input is first parsed into a structured query programming language using an LLM. Based on the parsed input, the *TalkToModel* generates the response by filling in pre-defined templates that define specific operations of the ML model. The responses from the templates are joined together into a final response that is shown to the user. The pre-defined templates ensure trustworthy responses and reduce hallucinations by the LLM (Yao et al., 2023). Nguyen et al. (2023) use LLM models to focus on the explanation of AI. They construct a question phrase bank and corresponding template-based answers using the information from the ML model. Therefore, the LLM is bound to specific rules and answering templates. The question is processed to a standard format where placeholders are filled with the information provided by the user. After matching the user input to the corresponding reference question and template, the relevant information from the prediction of the model needs to be retrieved. The retrieved SHAP value is used to explain the reason for a certain prediction output. In summary, to the best of the authors' knowledge, there are no other papers yet published that use the neural capabilities of a fine-tuned LLM to interpret AI and thereby offer a chatbot interaction for human understanding. The current state of the related work primarily focuses on pre-defined templates or question phrase banks that restrict the ability to adapt to complex, unforeseen scenarios or provide personalized explanations. By leveraging the LLM without these constraints, we enable the model to generate more dynamic, contextually rich responses that can better address the specific needs and nuances of each interaction, offering greater flexibility and in-depth explanations.

## 3 Research Design

We use DSR (Hevner et al., 2004; Peffers et al., 2007) to address our RQ of how an LLM can be leveraged to enhance existing XAI outputs for improved human understanding. Our research yields a reference architecture detailing the interaction of an XAI-evaluated ML model with a fine-tuned LLM model to enhance human interpretability. The reference architecture differentiates between the involved actors: user, system, and developer. Amongst the actors, activities and data, the reference architecture outlines the required information flows and interactions. Furthermore, the referential architecture is realized in a natural environment by providing a physical implementation in the form of a prototypical instantiation to evaluate the effectiveness of our proposed artifact and to iteratively improve the artifact in its design. For the design of our artifact, we follow the incremental and iterative DSR process of Peffers et al. (2007), which includes six phases: 1. problem identification, 2. definition of DOs, 3. design and development, 4. demonstration, 5. evaluation, and 6. communication. In the context of the problem identification, we identified and elaborated on the research problem to ensure that the objective of our DSR project aims for a relevant business problem (Hevner et al., 2004). Therefore, we derive DOs from the problem specification and an initial screening of the existing literature in the fields of XAI and LLMs. We structure the DSR project as a search process within the solution space defined by the DOs to enable the proper real-world instantiation of the prototype (Hevner et al., 2004; Peffers et al., 2007). We iteratively develop the abstract reference architecture detailing the interaction of an XAI-evaluated ML model with a fine-tuned LLM model by concretizing the DOs and initiating a real-world prototype on a use-case for battery health. We combine the demonstration and evaluation phase as we incorporate multiple evaluation activities into the development process (Hevner et al., 2004; Peffers et al., 2007) and adopt the FEDS framework to reduce uncertainty and risks during the design process and to underpin the effectiveness of the developed artifact (Venable et al., 2016). The FEDS framework structures design phases and evaluation episodes along the functional purpose and the context of the evaluation to ensure that the utility, quality, and efficacy of the artifact can be rigorously evaluated. Furthermore, we extend FEDS with selected components from the DSR evaluation framework by Sonnenberg and Vom Brocke (2012). In the context of FEDS, we follow the Technical Risk and Efficacy strategy as our intended artifact represents a technically oriented model requiring the artifact to meet the outlined research need through validation by real users in their artificial context. We split the evaluation strategy into three phases: 1. ex-ante evaluation, 2. intermediate evaluation, and 3. ex-post evaluation (Venable et al., 2016). The ex-ante evaluation is conducted immediately after the start of the research project in line with formative evaluations to improve the evaluation of the intended research outcome (Venable et al., 2016). Operationally, the evaluation phase is adapted from EVAL1 by Sonnenberg and Vom Brocke (2012). Thus, we surveyed the existing literature and conducted expert interviews to highlight the importance and novelty of our research and the defined DOs. The intermediate evaluation is an iterative procedure that aims to improve the expected applicability of the artifact and to provide initial insights into the artifact's behavior in a naturalistic context. Operationally, this evaluation phase combines guidelines from EVAL2 and EVAL3, as per Sonnenberg and Vom Brocke (2012) which involves a competing artifact analysis and initial demonstrations of the prototype to an audience, pertinent to the discipline. In the ex-post evaluation, we aimed to provide a summative assessment of the artifact's usefulness in different contexts by evaluating the impact of the research outcome (Venable et al., 2016). Operationally, the evaluation phase is adapted from EVAL4 by Sonnenberg and Vom Brocke (2012) and focuses on demonstrating the practical benefits of the artifact with real users. For this purpose, we conducted a survey with $n = 61$ participants in the context of a case study. To the best of our knowledge, this manuscript represents the first attempt to communicate the results of our research in the development and implementation of an interactive chatbot using fine-tuned LLMs for translating XAI results. The source code of our prototype is publicly available via the GitHub repository with the following link: https://github.com/bktllr/llm_explainable_ai

## 4     Design Specification

We define and present our derived DOs specifying the solution space and introduce the reference architecture, which uses a fine-tuned LLM to enhance human understanding of XAI. This serves as the core artifact of our research, incorporating the mentioned DOs.

### 4.1     Design Objectives

To guide the development of our artifact, we derive DOs as essential solution components based on the identification of the research problem and the screening of the existing literature (Peffers et al., 2007). The derived insights were synthesized into the DOs reflecting essential characteristics to solving the problem while aligning with established knowledge in that field. Our goal is to address the gap of personalized explanations of XAI (Schneider, 2024) while enhancing its interpretability for human users (Kaur et al., 2020). Below, we introduce and justify the derived DOs:

**DO1 – Understandable explanations improving the effort-benefit ratio for interpretation of XAI:**
Previous work on LLMs to enhance ML model interpretability highlights the importance of understandable explanations about how prevailing ML models work (Dhanorkar et al., 2021; Gao et al., 2022; Liao et al., 2020). Building on this foundation, our artifact aims to integrate generated textual descriptions of the ML model prediction with the corresponding XAI visualization. Thereby, humans should be enabled to interpret the latest predictions made by a prevailing ML model by reading through the chatbot-generated explanation while simultaneously being able to refer to the gained insights in a visual way. The combination of text and visual for human comprehension can increase cognitive load (Mayer & Moreno, 2003). However, research shows that this combination can also facilitate learning and engagement (Sweller et al., 2019). For example, chatbot-generated explanations can help users to better understand feature importances and contributions (positive vs. negative) to the ML model's output. This is particularly beneficial for users unfamiliar with XAI terminology, ensuring its broader accessibility and comprehension (Gao et al., 2022; Slack et al., 2023). While the combination of text and visualizations may require slightly more time for human interpretation compared to conventional approaches like standalone XAI visualizations, the interpretability of XAI should be enhanced significantly among different user groups – experts and non-experts in XAI.

**DO2 – Interactive and context-aware user experience facilitating personalized comprehension of XAI:**
Existing research underscores the significance of chatbots incorporating user feedback for personalized explanations and using natural language for interaction (Schneider & Handali, 2019; Schneider, 2024; Schwalbe & Finzel, 2023; Tenney et al., 2020). These capabilities are deemed essential for ensuring the continuous acceptance of a chatbot (Slack et al., 2023; Feine et al., 2020). To achieve this, the artifact should possess the ability to reference previous interactions by considering a chat history within the user interface. Additionally, to enable indirect user interaction with the generated predictions of the prevailing ML model, the chatbot should be able to refer to the existing XAI visualization within the user interface when generating responses. This should enable users to explore XAI visualizations interactively, aligning the human interpretation process with their existing knowledge in the field.

**DO3 – Inferential answer generation enabling the identification of new insights about the application domain of the prevailing ML model:**

While the aspect of inferential answer generation has been scarcely addressed in the underlying literature (Finch et al., 2021; Mazumder et al., 2018), it is crucial from the authors' perspective to consider it to ensure the usefulness of a chatbot for real-world applications. Therefore, the envisaged chatbot should not only be able to answer questions requiring general knowledge retrieved from the data it was trained on but also handle questions requiring inferences. These inferences should consider reasoning about specific ML model predictions, the prevailing XAI visual, and the context provided in the existing user chat, explaining why certain predictions were made. This approach aims to equip users with relevant insights that inspire solutions to real-world challenges.

## 4.2 Model Artifact

During the design and development phase, we iteratively specified and built the artifact, and its prototypical instantiation based on the DOs. Figure 1 provides a schematic visualization of the artifact.

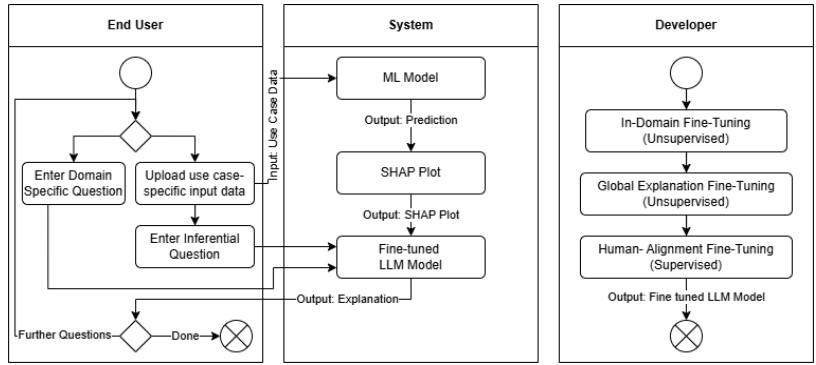

*Figure 1.        Our proposed artifact uses a fine-tuned LLM to enhance ML model interpretability visualized as a UML activity diagram (Dumas et al., 2001).*

Initially, the developer fine-tunes a chosen general-purpose LLM in three steps to improve its ability to process and respond to domain-specific content and thus, to meet the DOs. Within the application, the user first decides to utilize the chatbot either for domain-specific questions or for inferential questions. The system prompt includes information about expected questions, clarity and conciseness of answers, and the respective domain of the chatbot to deliver profound knowledge about a use case in a certain domain to the user. The user receives the answer from the fine-tuned LLM and can proceed with a new question to the system. In the case of inferential questions, the user is firstly required to upload use case-specific input data to the system to trigger a new ML prediction. In our use case, the ML prediction is conducted via a pre-trained CatBoost model (Prokhorenkova et al., 2018). The new prediction will then be analyzed automatically using an XAI technique to determine the corresponding importance score of SHAP. Both model prediction and XAI values are then integrated into an info prompt that will be automatically provided to the fine-tuned LLM, whenever the user submits a new question. The user receives the answer from the fine-tuned LLM on the chatbot interface and can proceed with a new question to the system. The presented artifact is unique in its ability to explain, interact with, and retrieve insights from ML models within a specific domain. This is achieved through the seamless integration of XAI techniques, ML models, and a fine-tuned LLM model. It can serve as a blueprint for system designs in other domains, requiring different XAI techniques, ML models, and fine-tuned LLM models.

## 4.3 Instantiation

### 4.3.1 Use Case

We implement and test a prototype of our artifact based on a use case for IoT battery health to increase the accessibility of the SoH. The SoH of a battery is a metric that indicates the current condition of the battery relative to its original state. It provides insights into how much the battery's capacity and performance have degraded over time (Bokstaller et al., 2023). Therefore, Battery Health Index (BHI) monitoring is a novel approach using a prediction model that enables continuous SoH monitoring based on decisive data sources (Bokstaller et al., 2024). The underlying prediction model is a pre-trained CatBoost model (Prokhorenkova et al., 2018) that predicts the SoH of IoT batteries based on battery-specific factors and environmental aspects, such as battery type, external influences, or usage behavior, which are represented in the features. Given the high relevance of BHI monitoring to ensure reliable battery operation, we utilize this pre-trained model to test the artifact regarding its capability to provide a meaningful explanation about the influence of individual model features and their interactions on the SoH of the battery.

### 4.3.2 Implementation

To implement the prototype, we opted for the *Llama-2-13b-chat-hf* (Touvron et al., 2023) as LLM which is part of the *Llama* series developed by Meta and provides a promising balance between quality and performance (Meta, 2023). This series represents state-of-the-art open-source models available at the time of this study, producing similar results as OpenAI's commercial GPT 3.5 (Touvron et al., 2023). To initialize our LLM in an accessible and interactive manner, we leverage the Gradio framework (Abid et al., 2019) to construct a chatbot interface that seamlessly integrates our fine-tuned Llama-2 in the backend. Also, we use SHAP as the XAI technique since we have categorical features.

### 4.3.3 Fine-Tuning

We fine-tune Llama-2 in three steps using the Oobabooga text-generation-webui (Oobabooga, 2024) which supports LoRa training, supervised fine-tuning, unsupervised fine-tuning, perplexity evaluation, and the calculation of the evaluation loss. The goal of the three fine-tuning steps is to first provide the model with broad, foundational knowledge, followed by narrowing the focus to more domain-specific and instance-specific knowledge, gradually refining the model's understanding. After each fine-tuning step, we evaluate the performance of the LLM on a quantitative basis using the perplexity score (Goldberg et al., 2017) and the loss metric. The three fine-tuning steps are:

**1. In-domain fine-tuning**[1] focuses on enriching the LLM's knowledge in a specific domain. This fine-tuning step is conducted in an unsupervised manner on unstructured data to encourage the base LLM to discover underlying patterns of the targeted domain. We specify the two domain categories "SHAP" and "batteries" and gather unstructured text files for these categories from the web through an extensive exploration on Google and Google Scholar. From the search results, we selected ten documents per category retrieved from websites, articles, and papers that contain important information and are cited frequently. We transformed the documents into text files, undergoing a cursory manual refinement. Within each category, a document was earmarked for evaluating the impact of this fine-tuning step on model performance using the perplexity score resulting in nine training documents and one evaluation document per category. The outcome of the fine-tuning process is the 1. *In-Domain fine-tuning* model that is evaluated once per category.

**2. Global explanation fine-tuning**[2] aims to enhance the LLM's use case-specific knowledge by considering general model findings derived from an extensive analysis of the SHAP values. This fine-tuning step is conducted in an unsupervised manner as well. Therefore, we analyzed the SHAP values

---

[1] Training-epochs: 3; Learning-rate: 0.0003; LoRA Rank: 8; LoRA Alpha: 16; Batch Size: 128; Micro Batch Size: 4

[2] Training-epochs: 3; Learning-rate: 0.0003; LoRA Rank: 8; LoRA Alpha: 16; Batch Size: 128; Micro Batch Size: 4

manually and summarized the findings in an unstructured text document that is then provided to the LLM. We conduct the following three SHAP analysis steps utilizing the SHAP library (Lundberg & Lee, 2017): a) SHAP summary plot with all available data to assess input features in terms of their impact on the SoH, b) SHAP dependency plot for the 15 most important input features to identify dependencies and influences between those features (Lundberg & Lee, 2017), c) SHAP summary plot per battery type to get battery specific information. While this fine-tuning step did not yield significant new improvements, it helped the LLM gain a deeper understanding of the SHAP concept and the battery SoH prediction process. For more information about the individual plots listed above, we refer to Lundberg and Lee (2017). The resulting information is summarized in an unstructured training document to perform *2. Global Explanation fine-tuning*.

**3. Human alignment fine-tuning**[3] aims at improving the LLM's capability to provide an explainable and understandable model explanation for inferential questions from the end user. This fine-tuning step is conducted in a supervised manner on structured data similar to future usage, so that the fine-tuned LLM generates outputs that align more precisely with the desired task objectives. In this case, the LLM is trained to accurately interpret SHAP values in relation to model predictions and specific user queries. To support this process, we automatically generate a structured JSON document for training that includes instructions, relevant contextual information related to the query – such as brief data descriptions, the 20 most important features with their values and SHAP values – and the correct output in alpaca format. (Touvron et al., 2023). Using the same approach, we generate an evaluation document (context Q&A) to have matching training and evaluation prompts. Finally, we tested against the evaluation document in a loss evaluation to assess the impact of *3. Human-Alignment fine-tuning*.

### 4.3.4 Fine-Tuning Results

Table 1 presents the quantitative evaluation results per evaluation document following the three fine-tuning steps in the form of an ablation study.

| Fine-Tuning Step | 1. SHAP In-Domain | 1. Battery In-Domain | 2. Global Explanation | 3. Human-Alignment |
|---|---|---|---|---|
| Test Document | Comprehensive guide into SHAP | Wikipedia battery SoH articles | Paraphrased summary of results | Context Q&A Dataset |
| Evaluation Type | Perplexity | Perplexity | Perplexity | Loss |
| Llama-2 (Baseline) | 5.76 | 8.94 | 11.44 | 2.56 |
| + In-domain fine-tuning | 5.32 (7.6%) | 7.91 (11.5%) | 9.94 | 2.31 |
| + Global Explanation | 5.32 | 7.93 | 9.81 (1.3%) | 2.31 |
| + Human-Alignment (Final Model) | 5.35 | 7.92 | 9.93 | 1.12 (51.5%) |

*Table 1.    Overview of Llama-2 fine-tuning steps for powering the battery SoH chatbot whereby (x.xx%) refers to the improvement in the evaluation metric in % due to fine-tuning.*

Table 1 shows that the in-domain fine-tuning leads to a significant improvement in the main evaluation documents per category. This demonstrates that fine-tuning is essential for achieving the level of accuracy and contextual understanding required in specific domains, which cannot be sufficiently attained through prompt engineering alone. In contrast, the global explanation fine-tuning has a minor impact on perplexity. One reason for this might be that the analysis of the SHAP values in the data preparation phase during the global explanation fine-tuning did not lead to significant new insights.

---

[3] Training-epochs: 20; Learning-rate: 0.0003; LoRA Rank: 8; LoRA Alpha: 16; Batch Size: 128; Micro Batch Size: 4

Consequently, all the findings discovered during global explanation fine-tuning were already transferred to the LLM during the in-domain fine-tuning, which can explain the positive spillover effects. Human alignment fine-tuning has the most substantial impact on the loss when evaluating its main evaluation document. However, Table 1 reveals a negligible increase in perplexity values for the previously fine-tuned LLMs.

## 5 Evaluation

For a rigorous evaluation of the intended artifact, we integrate FEDS into the DSR approach conducting three evaluation phases: an ex-ante evaluation, an intermediate evaluation, and an ex-post evaluation.

### 5.1 Ex-ante Evaluation

We conduct the ex-ante evaluation before the design and development phase to verify the significance of our research problem and the identified DOs (Venable et al., 2016). The RQ emerged from a practical industry problem. After identifying the problem and research gap from an initial screening of the literature on XAI and LLMs, we derived a set of DOs for the design of an artifact that aims to provide a solution for the given research problem. To assess the significance and relevance of both the RQ and DOs, we engaged in 16 expert interviews well-versed in ML and Information Systems (IS). The experts have between 2 and 15 years of experience in ML and 2 to 20 years of experience in IS. Each expert was interviewed individually, with sessions averaging 30 minutes. First. the experts were asked to rate the importance of the given RQ and the DOs using a 5-point Likert scale from 1 ('not important') to 5 ('extremely important'). Next, they ranked the importance of the DOs by allocating a budget of 10 points across them.

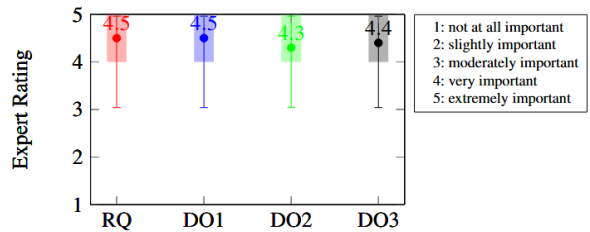

*Figure 2.      Ex-ante evaluation to rate the importance of the research question (RQ) and DOs.*

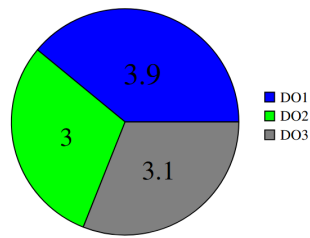

*Figure 3.      Ex-ante evaluation to rank the importance of the DOs.*

Figure 2 summarizes the importance ratings of the expert panel. The RQ was assessed as highly important with an average rating of 4.53 (out of 5), underscoring the significance of this research. Also, all DOs received a high importance rating of more than 4.4 (out of 5). Figure 3 summarizes the importance rankings of the expert panel. On average, the experts ranked DO1 as most important, followed by DO3 and DO2, with all DOs close to each other which is in line with the importance rating. The experts mentioned that it was difficult to rank the DOs as all are important. During the interview, they highlighted that both understandable explanations (DO1) and inferential answer generation (DO3) are core requirements for a chatbot design. For both DOs, experts emphasized that the need for their functionality varies depending on the user type. They noted that users with limited XAI knowledge, as well as those in decision-support roles, benefit more from these features compared to users with deeper XAI expertise and those primarily engaged in executing operational processes. Based on this feedback, we refined the DOs to clarify their intended impact. DO1 aims to enhance the efficiency of understanding XAI, particularly for users with limited XAI knowledge, while DO3 is designed to inspire insights into real-world solutions for those being involved in decision-support. Most experts expect the interactive experience (DO2) from a chatbot for concise understanding but see it often as a nice to have feature. However, the experts still mentioned the potential of clarifying follow-up questions on the XAI visualization and the personalized support by the LLM. Within the author team we decided that the aspects additionally mentioned by the experts like security or scalability of the application do not represent the core of the developed artifact and rather represent implementation-related problems. In summary, the results underscore the significance of our research, validating the DOs. We conclude that the ex-ante evaluation justifies our research problem and the DOs, being the foundation for the subsequent design of the artifact.

## 5.2 Intermediate Evaluation

The intermediate evaluation represents an iterative process that aims to gain knowledge about the behavior of the artifact with initial exposure to a naturalistic setup (Venable et al., 2016).

### 5.2.1 Competing Artifact Analysis

We conducted a competing artifact analysis by benchmarking the derived DOs of the artifact against the current state-of-the-art in the literature. More specifically, we identified whether existing works meet our DOs. The competing artifact analysis leads to several findings on the differentiation capabilities of our proposed artifact (Figure 1), which are summarized in Table 2.

| Approach | Prevailing artifact of this paper | Gao et al. (2022) | Nguyen et al. (2023) | Slack et al. (2023) |
|---|---|---|---|---|
| **DO1:** Understandable explanations improving the effort-benefit ratio for interpretation of XAI | Providing intuitive explanations of the underlying ML model supported by an XAI visual | Providing general explanations of the prevailing AI while giving indication about its confidence levels | Enabling natural language dialogues to understand XAI visualizations for a given ML model, e.g., by explaining feature importances | Enabling natural language dialogues to understand ML models through data-driven insights. |
| **DO2:** Interactive and context-aware user experience facilitating personalized comprehension of XAI | Facilitating ongoing human-chatbot conversations while referring to the generated XAI visual | Enabling human-chatbot interaction with limited flexibility due to a predefined conversation flow | Supporting continuous interaction through a chat history while grounding the chatbot to the generated XAI visualization | Supporting continuous interaction through a chat history while grounding the chatbot to the data used by the ML model and its predictions |
| **DO3:** Inferential answer generation enabling the identification of new insights about the application domain. | Connecting the derived insights from the XAI evaluation with domain-knowledge acquired within the LLM fine-tuning | Responses rely on pre-defined interactions | Responses rely on a constructed question phrase bank enabling template-based answers | Responses rely on pre-defined templates specifying the operations of the ML model |

*Table 2.        Competing artifact analysis*

In Table 2, a circle represents full, partial, or no consideration of the design objective. Therefore, our proposed artifact considers DO1 and DO2 to the full extent and DO3 partially. The instantiated prototype was able to draw inferences based on the generated XAI plots through the acquired domain knowledge within the in-domain fine-tuning. However, depending on the questions raised, the chatbot sometimes struggled with hallucination and the usage of domain-specific terms. Related work has made significant efforts to at least partially guarantee understandable explanations (DO1) and enable an interactive user experience (DO2). A core differentiation of our proposed artifact compared to prior works is the generation of inferential answers (DO3). No earlier paper included inferential answer generation based on the delivered data in the chatbot but relied on pre-defined templates or constructed question phrase banks. The competing artifact analysis shows the uniqueness of the design of our artifact in the current body of literature.

### 5.2.2  Demonstrations of the Prototype

We have iteratively developed the artifact, and the prototype based on feedback from weekly discussions in an explanatory focus group consisting of six members with interdisciplinary experience in XAI, LLM, ML, IoT battery technology, and IS. In addition, the prototype (shown in Figure 4) was demonstrated twice to a larger audience of about 20 people in a recurring meeting focused on sharing and discussing novel solutions in IS, ML and digital technologies. As a result, the participants brought an interdisciplinary background in these fields, fostering diverse perspectives on the prototype. The first demonstration of the prototype was conducted after the initial implementation of the pipeline, meaning that at this stage, we have initialized a general-purpose LLM without fine-tuning. We leveraged the Gradio framework to construct a chatbot interface that seamlessly integrates Llama-2 in the backend. The initial prototype showed fundamental XAI translation capabilities to facilitate the interpretation of ML models for human users and served as a first proof of the validity of the artifact. However, during the initial demonstration, the audience highlighted that the chatbot seems to struggle with detailed and transfer questions. Therefore, we prioritized further fine-tuning for the next development step.

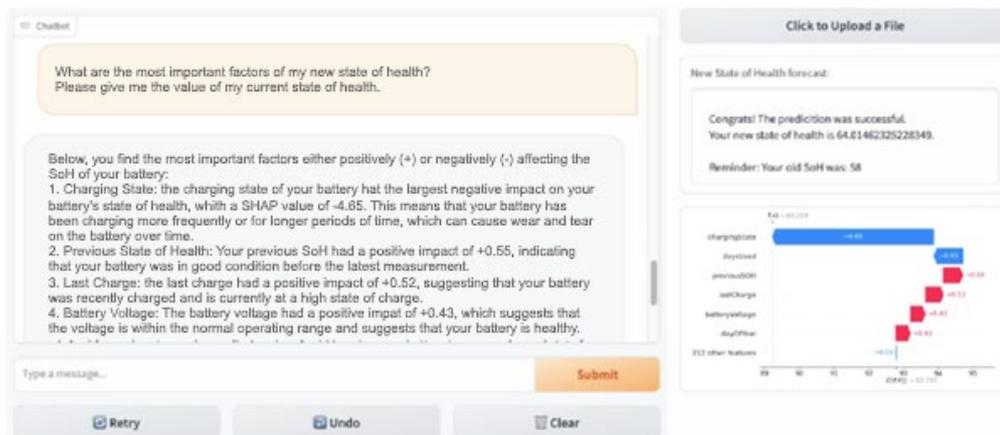

*Figure 4.*      *Snapshot of the chatbot interface.*

After the second demonstration of the prototype, the audience appreciated the inclusion of interactive features, such as the ability to upload battery data and allowing users to pose individual questions within the chatbot conversation. The chatbot's extensive domain-specific knowledge, acquired during fine-tuning, and its ability to demonstrate this knowledge when responding to user queries were highly commendable.

## 5.3      Ex-post Evaluation

To assess the effectiveness of our approach as an ex-post evaluation, we conducted an online survey. We aimed to measure the overall interpretability and clarity when interacting with our chatbot and to understand the variation in comprehensibility across participants with different proficiency levels. We reached out to a diverse sample of over 100 participants, varying in their familiarity with ML concepts, particularly SHAP values. We recruited participants by sending out a survey link with no incentive to individuals and teams in the industry, which we selected based on their professional position and knowledge in AI (low and high), and to different social and student groups to ensure diverse representation. From this pool, we received 61 responses, which formed the basis of our analysis. The survey was structured into four main sections: demographic information, evaluation of SHAP interpretability, assessment of explanations provided by a non-fine-tuned chatbot, and evaluation of our fine-tuned XAI chatbot in terms of clarity, interpretability, and cognitive effort. First, we wanted to gauge the proficiency of the user regarding ML and SHAP values. We did this by asking questions such as *"How familiar are you with AI concepts"* and *"How often do you interact with AI technologies in your daily life?"*. This classification helped to distinguish the users. Participants first interacted with a SHAP waterfall plot of a battery SoH prediction, accompanied by a basic explanation of the SHAP values and features (Baseline 1). They then responded to questions assessing the clarity, interpretability, and cognitive effort required to understand the plot. The same evaluation procedure was applied to a screenshot of explanations generated by a non-fine-tuned chatbot (Baseline 2), followed by our proposed fine-tuned XAI chatbot, which translates the same SHAP values into more user-friendly text. To minimize order effects, all participants first evaluated the SHAP plot, followed by the chatbot-generated explanations (Doshi-Velez & Kim, 2017). To evaluate the performance of our chatbot, we compared both baselines with our proposed fine-tuned XAI chatbot. Our findings, illustrated in Figure 5, indicate that our solution surpassed the conventional SHAP waterfall plot and non-fine-tuned chatbot across all three evaluation metrics (higher value means better result). Participants, regardless of proficiency levels, favored the user-friendly explanations provided by the fine-tuned XAI chatbot over the SHAP-generated plot and non-fine-tuned chatbot due to the lack of domain knowledge.

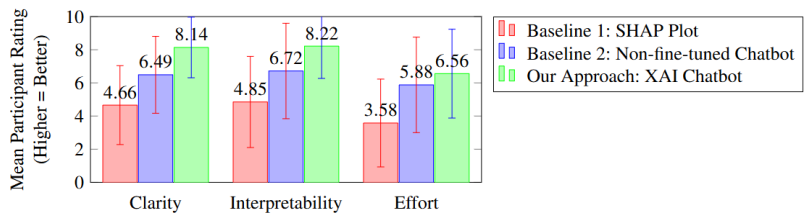

*Figure 5.* Survey results for overall interpretability and clarity.

Further, we compared the ratings of the two baselines against our fine-tuned XAI chatbot across participant proficiency levels (low, medium, and high). We sourced the proficiency levels from the demographic section of our survey. For each proficiency level, we compute the disparity between the baseline values and our chatbot across three specified categories. A positive disparity indicates a preference for our solution while a negative disparity suggests greater usability of the baseline solution. As shown in Figure 6, our solution consistently outperforms both baselines across all proficiency levels. However, it is noteworthy that as a participant's proficiency increases, the disparity over all three evaluation categories decreases. This trend can be attributed to the enhanced analytical skills and familiarity with SHAP plots among more proficient users. Users with lower proficiency levels encounter challenges in interpreting SHAP plots effectively. This is where our solution has a clear advantage. The chatbot can translate the SHAP plot into text that is also understandable for participants with low proficiency. Consequently, our analysis confirmed that participants with lower proficiency levels exhibited a stronger preference for our chatbot compared to those with higher proficiency levels.

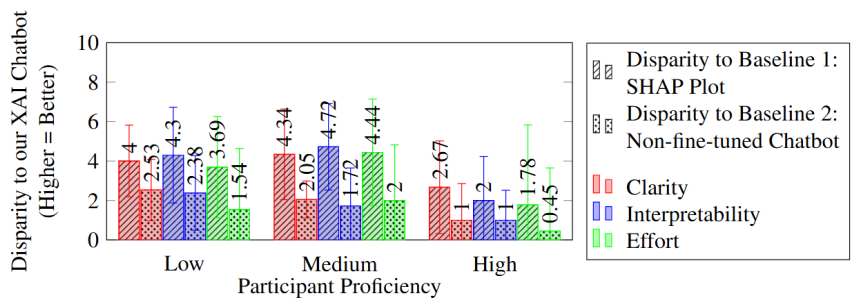

*Figure 6.* Survey results for comprehensibility disparities across proficiency levels.

# 6      Conclusion

Following a DSR approach, this research develops and presents a novel artifact to address the challenge of interpreting ML models. With the increasing complexity of ML models often seen as "black boxes" by end users, our solution enhances interpretability by combining XAI with fine-tuned LLMs, enabling users to query and understand model decisions through natural language interactions. This is especially crucial for various real-world applications requiring the safe operation of ML models as interpretability can contribute to assessing whether requirements such as transparency, fairness, data privacy, reliability, robustness, and trust, amongst others are fulfilled (Doshi-Velez & Kim, 2017). Our empirical survey findings underline significant enhancements in interpretability, particularly for users less familiar with XAI. By embedding SHAP into a conversational interface, we have successfully translated complex SHAP plots into high-level, easy-to-understand insights, thus lowering the knowledge requirements for end users. Thereby, this research contributes significantly to the advancements of XAI by bridging the gap between complex ML models and human comprehension. Our solution achieves this by delivering understandable explanations, offering an interactive user experience, and responding to domain-specific questions and inferential questions based on the prediction of an ML model. Unlike previous works, we are not using template-based answer and response matching but rather apply a fine-tuned LLM based on SHAP values as input. To the best of the writers' knowledge, we are the first paper presenting the fine-tuning steps on how to adapt the LLM to work with SHAP values and publish the corresponding code. While we demonstrate one prototypically instantiation of our artifact for a battery SoH use case using SHAP values, its potential extends far beyond this specific application. The versatile design of our artifact allows it to be generalized across various other domains, significantly enhancing trust, transparency, and reliability (Javed et al., 2023). For example, in smart city solutions, XAI can improve transparency in areas such as flood detection, drainage monitoring (Thakker et al., 2020), and accident prevention (Lavrenovs & Graf, 2021). In the medical field, the ability to provide clear and interpretable explanations for machine decisions and predictions is essential for ensuring their reliability and building trust (Tjoa & Guan, 2021). Therefore, having an adaptive and generalizable artifact that can be applied across various sectors is of great importance. Future work can implement and extend our artifact to such use cases and different XAI techniques to investigate its performance. Further extensions of the artifact could include the integration of unsupervised ML models and distinct techniques to adapt the LLM beyond the presented fine-tuning techniques. Our LLM is fine-tuned to domain specific knowledge regarding batteries and SHAP, providing a transferable approach for use-cases in these domains. However, for other application domains with limited common knowledge, an LLM may require more extensive fine-tuning to achieve effective inferential reasoning, particularly in contrast to well-established fields where much of the relevant knowledge is already embedded in the pre-trained model. Since acquiring sufficient domain-specific data for fine-tuning can be challenging, an alternative or complementary approach could be the integration of Retrieval-Augmented Generation (RAG). By incorporating relevant external information, RAG can mitigate scenarios where LLM hallucinations occur and thereby can reduce instances of incorrect information generation. Our study comes with the following limitations: First, our artifact is built upon and evaluated based on current technology and their inner working, such as SHAP, ML, and LLM, and thus, limited to their current capabilities. This study uses SHAP as a well-established XAI technique providing ex-post explanations for ML models to enhance model interpretability. Second, to ensure not only transparent but also trustworthy ML models, extending our artifact to integrate prediction confidence scores could be a promising research extension. Third, gaining insights into users' evolving needs requires a longer-term evaluation. While the current assessment captures a snapshot through interviews and an online survey with a broader audience, extending the evaluation over time could provide deeper insights into user experience and adaptation.